%% file: emnlp2021.tex
\pdfoutput=1

\documentclass[11pt]{article}

\usepackage{emnlp2021}

\usepackage{times}
\usepackage{latexsym}

\usepackage[T1]{fontenc}

\usepackage[utf8]{inputenc}

\usepackage{microtype}

\usepackage{bm}
\usepackage{times}
\usepackage{latexsym}
\usepackage{amsmath,amsfonts,amssymb,bbm,epsfig,bm}
\usepackage{balance}
\usepackage{booktabs}
\usepackage{multirow}
\usepackage{array}
\usepackage{url}
\usepackage{tikz}
\usepackage{makecell}
\usepackage{subfigure}
\usepackage{framed}
\usepackage{pstricks}

\usepackage{graphicx}
\usepackage{xcolor}
\usepackage{color}
\usepackage{enumerate}
\usepackage{arydshln}
\usepackage{algorithm}
\usepackage{algpseudocode} 
\usepackage{pifont}
\usepackage{xspace}

\usepackage{nicefrac}
\usepackage{amsthm}
\usepackage{mathtools}
\usetikzlibrary{arrows}
\usepackage{enumitem}
\usepackage{hyperref}
\usepackage{caption}

\newtheorem{thm}{Theorem}[]
\newtheorem{thm1}{Theorem}[]
\newtheorem{lem}{Lemma}[]

\newcommand{\vect}[1]{\mathbf{#1}}

\renewcommand{\algorithmiccomment}[1]{\bgroup\hfill $\triangleright$ ~#1\egroup}

\newcommand{\ChildTuning}{\textsc{Child-Tuning}\xspace}
\newcommand{\ChildTuningF}{\textsc{Child-Tuning}$_F$\xspace}
\newcommand{\ChildTuningD}{\textsc{Child-Tuning}$_D$\xspace}
\newcommand{\CTuningF}{\textsc{C.Tuning}$_F$\xspace}
\newcommand{\CTuningD}{\textsc{C.Tuning}$_D$\xspace}

\newcommand{\mycolor}[1]{\textcolor[RGB]{64,101,149}{#1}}
\newcommand{\mydarkcolor}[1]{\textcolor[RGB]{64,101,149}{#1}}

\title{Raise a Child in Large Language Model: \\ Towards Effective and Generalizable Fine-tuning}

\author{First Author \\
  Affiliation / Address line 1 \\
  Affiliation / Address line 2 \\
  Affiliation / Address line 3 \\
  \texttt{email@domain} \\\And
  Second Author \\
  Affiliation / Address line 1 \\
  Affiliation / Address line 2 \\
  Affiliation / Address line 3 \\
  \texttt{email@domain} \\}

\author{
  Runxin Xu$^{1}$\footnotemark[1], 
  Fuli Luo$^{2}$\thanks{\, Equal Contribution. Joint work between Alibaba and Peking University.}, 
  Zhiyuan Zhang$^{1}$, 
  Chuanqi Tan$^{2}$, \\
  \bf Baobao Chang$^{1}$\footnotemark[2],
  \bf Songfang Huang$^{2}$\thanks{\, Corresponding authors.},
  \bf Fei Huang$^{2}$
  \\
  $^{1}$Key Laboratory of Computational Linguistics, Peking University, MOE, China\\
  $^{2}$Alibaba Group\\
  \texttt{
     runxinxu@gmail.com, \{zzy1210,chbb\}@pku.edu.cn
  } \\
  \texttt{
     \{lfl259702,chuanqi.tcq,songfang.hsf,f.huang\}@alibaba-inc.com
  }
}

\begin{document}
\maketitle

\begin{abstract}
\input{main/abstract.tex}
\end{abstract}

\section{Introduction}
\input{main/intro.tex}

\section{Methodology}
\input{main/method.tex}

\section{Experiments}

\input{main/experiments.tex}

\section{Analysis and Discussion}
\input{main/discussion}

\section{Related Work}
\input{main/related.tex}

\section{Conclusion}
\input{main/conclusion.tex}

\section*{Acknowledgments}
This paper is supported by the National Key Research and Development Program of China under Grant No. 2020AAA0106700, the National Science Foundation of China under Grant No.61936012 and 61876004.

\bibliography{custom}
\bibliographystyle{acl_natbib}

\appendix

\input{main/appendix}
\input{main/appendix-theory}

\end{document}

%% file: main/abstract.tex
Recent pretrained language models extend from millions to billions of parameters. Thus the need to fine-tune an extremely large pretrained model with a limited training corpus arises in various downstream tasks.
In this paper, we propose a straightforward yet effective fine-tuning technique, \ChildTuning, which updates a subset of parameters (called child network) of large pretrained models via strategically masking out the gradients of the non-child network during the \textit{backward} process.
Experiments on various downstream tasks in GLUE benchmark show that \ChildTuning consistently outperforms the vanilla fine-tuning by $1.5\sim8.6$ average score among four different pretrained models, and surpasses the prior fine-tuning techniques by $0.6\sim1.3$ points. 
Furthermore, empirical results on domain transfer and task transfer show that \ChildTuning can obtain better generalization performance by large margins.

%% file: main/intro.tex
Pretrained Language Models (PLMs) have had a remarkable effect on the natural language processing (NLP) landscape recently~\citep{bert,roberta,electra}. Pretraining and fine-tuning have become a new paradigm of NLP, dominating a large variety of tasks.

Despite its great success, how to adapt such large-scale pretrained language models with millions to billions of parameters to various scenarios, especially when the training data is limited, is still challenging.
Due to the extremely large capacity and limited labeled data, conventional transfer learning tends to \textit{aggressive} fine-tuning~\citep{smart}, resulting in: 1) degenerated results on the test data due to overfitting~\citep{bert,Phang2018,mixout}, and 2) poor generalization ability in transferring to out-of-domain data or other related tasks~\citep{VIB,rxf}.

\input{float/figure-model}

Preventing the fine-tuned models to deviate too much from the pretrained weights (i.e., with less knowledge forgetting), is proved to be effective to mitigate the above challenges~\citep{distance}. 
For instance, RecAdam~\citep{RecAdam} introduces $L_2$ distance penalty between the fine-tuned weights and their pretrained weights.
In addition, Mixout~\citep{mixout} randomly replaces part of the model parameters with their pretrained weights during fine-tuning. The core idea behind them is to utilize the pretrained weights to regularize the fine-tuned model.  

In this paper, we propose to mitigate the aggressive fine-tuning problem from a new perspective. 
Based on the observation that it is unnecessary to update all the parameters within the large-scale model during fine-tuning, we propose an effective fine-tuning technique, \textit{\ChildTuning}, which straightforwardly updates a subset of parameters (called \emph{child} network) via strategically \textit{\textbf{masking}} out the gradients of non-child network in the backward process, as illustrated in Figure~\ref{fig:model}. Note that it is different from model pruning, since it still forwards on the whole network, thus making the full use of knowledge hidden in the pretrained weights.

In detail, we propose two variants, \ChildTuningF and \ChildTuningD, which respectively detect the child network in a \textbf{task-free} and a \textbf{task-driven} way.
\ChildTuningF chooses out the child network in the absence of task data via a Bernoulli distribution.
It introduces noise to the full gradients, playing a role of regularization, hence preventing overfitting to small datasets and leading to better generalization.
Furthermore, \ChildTuningD utilizes the downstream task data to detect the most task-related parameters as the child network and freezes the parameters in non-child network to their pretrained weights.
It decreases the hypothesis space of the model via a task-specific mask applied to the full gradients, helping to effectively adapt the large-scale pretrained model to various tasks and meanwhile greatly maintain its original generalization ability.

Our extensive experiments on the GLUE benchmark show that \ChildTuning can be more excellent at fine-tuning different PLMs, with up to $8.60$ average score improvement on CoLA/RTE/MRPC/STS-B tasks compared to  vanilla fine-tuning (Section.~\ref{sec:main-results}). Moreover, it achieves better generalization ability in transferring to out-of-domain data and other related tasks (Section.~\ref{sec:generalization}).
Experimental results also demonstrate that \ChildTuning yields consistently greater improvements than state-of-the-art fine-tuning methods.
More importantly, since \ChildTuning is orthogonal to these prior methods, integrating \ChildTuning with them can even lead to further improvements (Section.~\ref{sec:other-methods}).

In summary, our contributions are three-fold:
\begin{itemize}
    \item We propose \ChildTuning, a straightforward yet effective fine-tuning technique that only updates the parameters in the child network. We explore to detect the child network in both task-free and task-driven ways.
    \item \ChildTuning can effectively adapt the large-scale pretrained model to various downstream scenarios, from in-domain to out-of-domain, and cross-task transfer learning.
    \item Since \ChildTuning is orthogonal to prior fine-tuning methods, integrating \ChildTuning with them can further boost the fine-tuning performance.
\end{itemize}

%% file: float/figure-model.tex
\begin{figure}[t]
	\centering
    \includegraphics[width=0.48\textwidth]{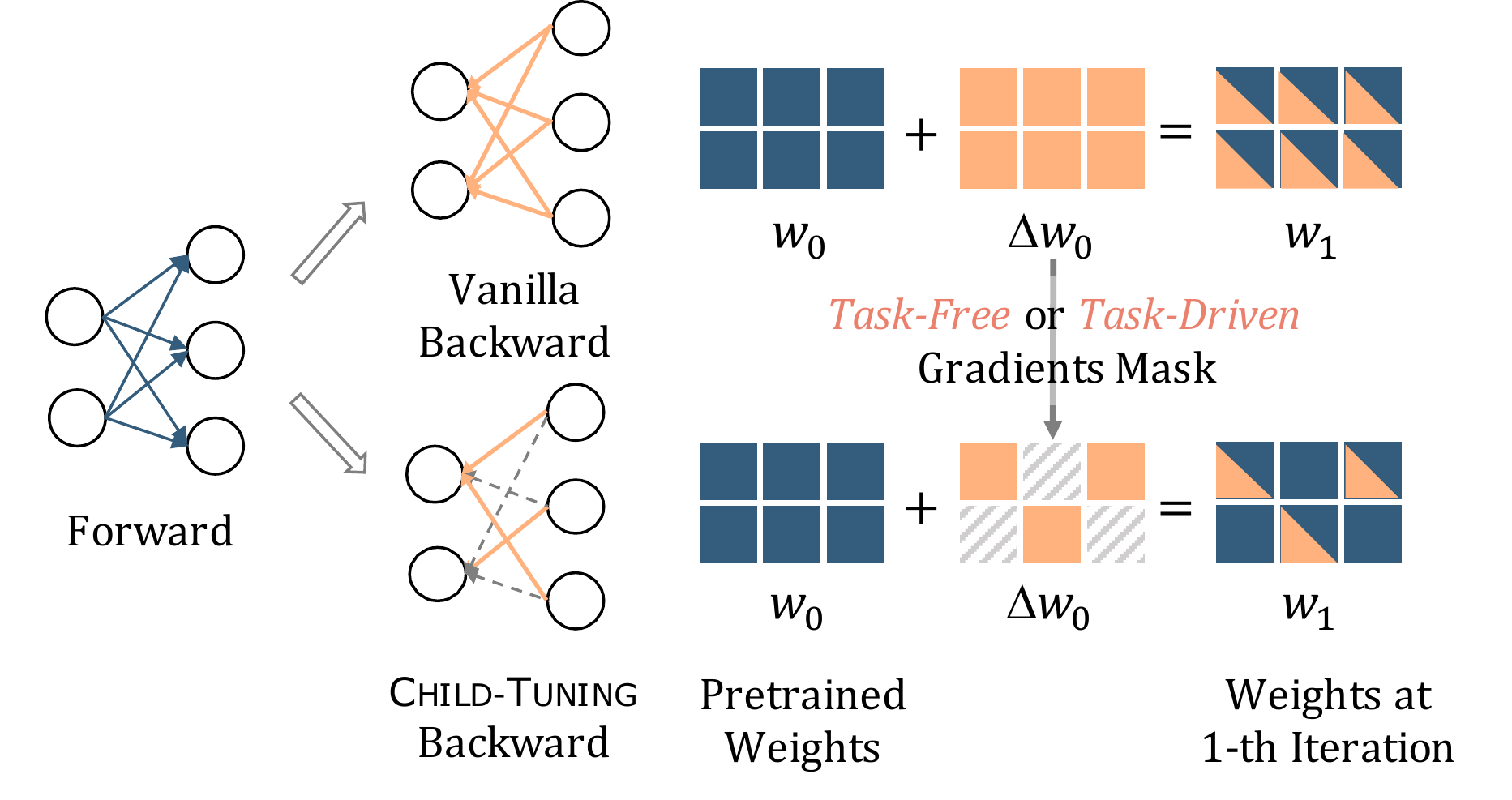}
	\caption{The illustration of \ChildTuning.
	\underline{\textit{Left}}: It forwards on the whole network while backwarding on a subset of network (i.e., child network).
    \underline{\textit{Right}}: To achieve this, a task-free or task-driven mask is performed on the gradients of the non-child network, resetting them to zero (grey diagonal grids).
	}
	\label{fig:model}
\end{figure}

%% file: main/method.tex
To better adapt large-scale pretrained language model to various downstream tasks, we propose a simple yet effective fine-tuning technique, \ChildTuning.
We firstly introduce a gradient mask in the backward process to achieve the aim of updating a subset of parameters (i.e., child network),  while still utilizing the knowledge of the whole large model in the forward process (Section~\ref{sec:overview}).
Then, we explore two ways to detect the child network (i.e., generate different gradient masks): \ChildTuningF that are in a task-free way (Section~\ref{sec:task-free}), and \ChildTuningD that are in a task-driven way ( Section~\ref{sec:task-driven}).

\subsection{Overview of \ChildTuning}
\label{sec:overview}
We start the introduction of \ChildTuning by giving a general formulation of the back propagation during the vanilla fine-tuning. We denote the parameters of the model at the $t$-th iteration as $\vect{w}_{t}$ ($\vect{w}_0$ refers to the pretrained weights). The vanilla fine-tuning computes the gradient of the loss $\mathcal{L}(\vect{w}_{t})$ and then applies gradient descent to all parameters, which can be formulated as:
\begin{align}
\vect{w}_{t+1} & = \vect{w}_{t} - \eta \frac{\partial \mathcal{L}(\vect{w}_{t})}{\partial \vect{w}_{t}}
\label{eq:delta-w}
\end{align}
where $\frac{\partial \mathcal{L}(\vect{w}_{t})}{\partial \vect{w}_{t}}$ are the gradients corresponding to the model parameters $\vect{w}_{t}$, $\eta$ is the learning rate.

\ChildTuning also backwardly computes the gradients of all trainable parameters like standard fine-tuning.
However, the key difference is that \ChildTuning determines a child network $\mathcal{C}_t$ at the $t$-th iteration, and only updates this part of parameters.
To achieve this, we firstly define a $0$-$1$ mask that is the same-sized as $\vect{w}$ as follows:
\begin{equation}
\mydarkcolor{\bm{M}^{(i)}_{t}} =\left\{
\begin{array}{lc}
1 , & \vect{w}^{(i)}_{t} \in \mathcal{C}_t\\
0 , & \vect{w}^{(i)}_{t} \notin \mathcal{C}_t
\end{array} \right.
\end{equation}
where $\bm{M}^{(i)}_{t}$ and $\vect{w}^{(i)}_{t}$ denote the $i$-th element of the mask $\bm{M}_{t}$ and parameters $\vect{w}_{t}$ at the $t$-th training iteration, respectively.

Then, we formally define \ChildTuning technique by simply replacing Eq.~\ref{eq:delta-w} with the following equation:
\begin{equation}
    \vect{w}_{t+1} = \vect{w}_{t} - \eta \frac{\partial \mathcal{L}(\vect{w}_{t})}{\partial \vect{w}_{t}} \odot  \mydarkcolor{\bm{M}_{t}}
  \label{eq:backward}
\end{equation}

\input{alg/alg-new}

Algorithm~\ref{algo:adam} provides the pseudo-code of \ChildTuning when applied to widely used Adam~\citep{adam} optimizer.
The main difference is the insertion of line $5$-$7$. 

\subsection{Task-Free Variant: \ChildTuningF} \label{sec:task-free}
In this section, we firstly explore the choice of the child network that does not require any downstream task data, i.e., a task-free technique called \ChildTuningF.
Specifically, \ChildTuningF generates a $0$-$1$ mask $\bm{M_t}$ at the $t$-th iteration drawn from a Bernoulli distribution with a probability $p_F$:
\begin{equation}
\bm M_{t} \sim \mathrm{Bernoulli}(p_F)
\end{equation}

The higher the $p_F$ is, the larger the child network is, and hence more parameters are updated.
When $p_F=1$, \ChildTuningF degenerates into the vanilla fine-tuning method.
Note that we also enlarge the reserved gradients by $\frac{1}{p_F}$ to maintain the expectation of the gradients.

We theoretically justify the effectiveness of \ChildTuningF.
We denote $\vect{\Delta w}$ as the update at each iteration:
\begin{align}
    \vect{\Delta w} = \eta \frac{\partial \mathcal{L}(\vect{w})}{\partial \vect{w}} \odot  \bm{M}
\end{align}
Intuitively, Theorem~\ref{thm:mainbody-childtuningf} shows the variance of gradients is a strictly decreasing function of $p_F$.
Thus, \ChildTuningF improves the variance of the gradients, and the trade-off between exploration and exploitation can be controlled by adjusting $p_F$.
As illustrated in Theorem~\ref{thm:mainbody-childtuningf2}, with higher variance, the model can converge to more flat local minima (smaller $\rho$ in Theorem~\ref{thm:mainbody-childtuningf2}).
Inspired by studies that show flat minima tends to generalize better~\citep{KeskarMNST17,abs-2006-05620, foret2021sharpnessaware}, we can further prove \ChildTuningF decreases the generalization error bound.

\begin{thm}
\label{thm:mainbody-childtuningf}
Suppose $\mathcal{L}$ denotes the loss function on the parameter $\vect{w}$, 
the gradients obey a Gaussian distribution $\mathcal{N}(\frac{\partial \mathcal{L}}{\partial \vect{w}},\sigma^2_\vect{g}\mathbf{I}_k)$, 
and SGD with learning rate $\eta$ is used.
For a randomly sampled batch $\mathcal{B}$, 
if GradMask reserves gradients with probability $p_F$, the mean and covariance of the update $\vect{\Delta w}$ are,
\begin{align}
\mathbb{E}[\vect{\Delta w}]&=-\eta \frac{\partial \mathcal{L}}{\partial \vect{w}}\\
\Sigma[\vect{\Delta w}]&=\frac{\eta^2\sigma_\vect{g}^2\mathbf{I}_k}{p_F|\mathcal{B}|}+
\frac{(1-p_F)\eta^2\text{diag}\{\frac{\partial \mathcal{L}}{\partial \vect{w}} \}^2}{p_F}
\end{align}

Specially, when $\vect{w}$ is a local minima, $\mathbb{E}[\vect{\Delta w}]=\vect{0}_k, \Sigma[\vect{\Delta w}]=\sigma^2\mathbf{I}_k$ and $\sigma^2=\frac{\eta^2\sigma_\vect{g}^2}{p_F|\mathcal{B}|}$ is a strictly decreasing function of $p_F$.
\end{thm}

\begin{thm}
\label{thm:mainbody-childtuningf2}
Suppose $\vect{w}_0$ denotes the pretrained parameter; $k$ is the number of parameters; $\vect{w}$ denotes the local minima the algorithm converges to; $\rho$ is the greatest eigenvalue of the Hessian matrix on $\vect{w}$, which indicates the sharpness.
If $\vect{\Delta w}\sim N(\vect{0}_k, \sigma^2\mathbf{I}_k)$, when the following bound holds, the algorithm can converge to the local minima $\vect{w}$ with high probability,
\begin{align}
\rho\le O\big(\frac{1}{\sigma^2}\big)
\end{align} 

Suppose the prior over parameters after training is $P=N(\vect{w}_0, \sigma_0^2\mathbf{I}_k)$, the following generalization error bound holds with high probability,
\begin{align}
\resizebox{0.42\textwidth}{!}{
$\text{bound}(\vect{w}) \le O\big(\frac{k\sigma_0^2-\|\vect{w}-\vect{w}_0\|^2}{\sigma^2}\big)+\mathcal{R}$
}
\end{align} 
where $\mathcal{R}$ is a term not determined by $\sigma$.
\end{thm}

Thus, \ChildTuningF can be viewed as a strong regularization for the optimization process.
It enables the model to skip the saddle point in the loss landscape and encourages the model to converge to a more flat local minima.
Please refer to Appendix~\ref{sec:theory} for more details about stated theorems and proofs.

\subsection{Task-Driven Variant: \ChildTuningD} \label{sec:task-driven}

Taking the downstream labeled data into consideration, we propose \ChildTuningD, which detects the most important child network for the target task.
Specifically, we adopt the Fisher information estimation to find the highly relevant subset of the parameters for a specific downstream task. 
Fisher information serves as a good way to provide an estimation of how much information a random variable carries about a parameter of the distribution~\citep{7560179,7472157}.
For a pretrained model, Fisher information can be used to measure the relative importance of the parameters in the network towards the downstream tasks.

Formally, the Fisher Information Matrix (FIM) for the model parameters $\vect{w}$ is defined as follows:
\begin{align*}
\resizebox{0.98\linewidth}{!}{$
    \vect{F}\left (\vect{w} \right ) = \mathbb{E} \left [(\frac{\partial \log p\left (y|\vect{x};\vect{w} \right )}{\partial \vect{w}})(\frac{\partial \log p\left (y|\vect{x};\vect{w} \right )}{\partial \vect{w}})^\top \right ]$}
\end{align*}
where $\vect{x}$ and $y$ denote the input and the output respectively.
It can be also viewed as the covariance of the gradient of the log likelihood with respect to the parameters $\vect{w}$. 
Following~\citet{EWC}, given the task-specific training data data $D$, we use the diagonal elements of the empirical FIM to point-estimate the task-related importance of the parameters.
Formally, we derive the Fisher information for the $i$-th parameter as follows:
\begin{align}
    \vect{F}^{(i)}\left (\vect{w} \right ) = \frac{1}{\left|D\right|} \sum_{j=1}^{\left| D \right |} 
    \left (\frac{\partial \log p \left (y_j|\vect{x}_j;\vect{w} \right )}{\partial \vect{w}^{(i)}} \right )^2
\label{eq:fisher}
\end{align}

We assume that the more important the parameter towards the target task, the higher Fisher information it conveys.
Hence the child network $\mathcal{C}$ is comprised of the parameters with the highest information. The child network ratio is $p_D=\frac{\left | \mathcal{C} \right | }{\left | \mathcal{C} \right | + \left | \overline{\mathcal{C}} \right | } \in (0, 1]$, where $\overline{\mathcal{C}}$ denotes the non-child network.
As $p_D$ rises, the scale of the child network also increases, and when $p_D=1$ it degenerates into the vanilla fine-tuning strategy.

\input{float/table-main}

Since the overhead of obtaining the task-driven child network is heavier than that of the task-free one, we simply derive the child network for \ChildTuningD at the beginning of fine-tuning, and keep it unchanged during the fine-tuning, i.e., $\mathcal{C}_0=\mathcal{C}_1=\dots=\mathcal{C}_T$.
In this way, \ChildTuningD dramatically decreases the hypothesis space of the large-scale models, thus alleviating overfitting. Meanwhile, keeping the non-child network freezed to their pretrained weights can substantially maintain the generalization ability.

%% file: alg/alg-new.tex
\algnewcommand{\LineComment}[1]{\Statex ~~~~~~\textsc{//}~\textit{#1}}

\newlength{\textfloatsepsave} 
\setlength{\textfloatsepsave}{\textfloatsep} 
\setlength{\textfloatsep}{-30pt}
\begin{algorithm}[t]
\caption{\ChildTuning for Adam Optimizer}
\label{algo:adam}
\begin{algorithmic}[1]
\Require
$\vect{w}_0$: initial pretrained weights;
$\mathcal{L}(\vect{w})$: stochastic objective function with parameters $\vect{w}$;
$\eta$: learning rate;
$\beta_1,\beta_2\in [0,1)$: exponential decay rates for the moment estimates;
\State{\textbf{initialize} timestep $t \leftarrow 0$, first moment vector $m_{0} \leftarrow 0$, second moment vector $v_{0} \leftarrow 0$}
\While{not converged}\label{algo:adam:iteration}
\State $t \gets t + 1$
\LineComment{Get gradients}
\State $\vect{g}_t \gets \frac{\partial \mathcal{L}(\vect{w}_{t})}{\partial \vect{w}_t} $ 
\mydarkcolor{\LineComment{Get task-free/task-driven child network}}
\State $\mathcal{C}_t \gets \text{GetChildNetwork}()$ 
\mydarkcolor{\LineComment{Generate a corresponding gradient mask}}
\State $\bm{M}_t \gets \text{GenerateMask}(\mathcal{C}_t)$ 
\mydarkcolor{\LineComment{Employ mask for gradients}}
\State $\vect{g}_t \gets \vect{g}_t \odot \bm{M}_t$ 
\State $\vect{m}_t \gets \beta_1 \cdot \vect{m}_{t-1} + (1-\beta_1) \cdot \vect{g}_t$ 
\State $\vect{v}_t \gets \beta_2 \cdot \vect{v}_{t-1} + (1-\beta_2) \cdot \vect{g}^2_t$ 
\LineComment{Bias correction} \label{algo:adam:bias-first}
\State $\hat{\vect m}_t \gets \vect{m}_t / (1-\beta_1^t)$
\State $\hat{\vect v}_t \gets \vect{v}_t / (1-\beta_2^t)$ 
\LineComment{Update weights} \label{algo:adam:update}
\State $\vect{w}_t \gets \vect{w}_{t-1} - \eta \cdot \hat{\bm m}_t / (\sqrt{\hat{\bm v}_t} + \epsilon)$
\EndWhile
\State \Return $\vect{w}_t$
\end{algorithmic}
\end{algorithm}
\setlength{\textfloatsep}{\textfloatsepsave}

%% file: float/table-main.tex
\begin{table*}[t]
\centering
\scalebox{0.93}{
    \begin{tabular}{lcccccccccc}
    \toprule
    \multirow{2}{*}{\bf Method} & \multicolumn{5}{c}{BERT} & 
    \multicolumn{5}{c}{XLNet} \\
    \cmidrule(lr){2-6} \cmidrule(lr){7-11}
    ~ & CoLA & RTE & MRPC & STS-B & \underline{Avg} & CoLA & RTE & MRPC & STS-B & \underline{Avg} \\
    \midrule
    Vanilla Fine-tuning & 63.13 & 70.18 & 90.77 & 89.61 & \underline{78.42} & 47.14 & 77.62 & 91.90 & 91.77 & \underline{77.11} \\
    \ChildTuningF & 63.71 & 72.06 & 91.22 & \bf 90.18 & \underline{79.29} & \bf 52.07 & 78.05 & 92.29 & 91.81 & \underline{78.56} \\
    \ChildTuningD & \bf 64.92 & \bf 73.14 & \bf 91.42 & \bf 90.18 & \bf \underline{79.92} & 51.54 & \bf 80.94 & \bf 92.46 & \bf 91.82 & \bf \underline{79.19} \\
    \midrule
    \multirow{2}{*}{\bf Method} & \multicolumn{5}{c}{RoBERTa} & 
    \multicolumn{5}{c}{ELECTRA} \\
    \cmidrule(lr){2-6} \cmidrule(lr){7-11}
    ~ & CoLA & RTE & MRPC & STS-B & \underline{Avg} & CoLA & RTE & MRPC & STS-B & \underline{Avg}  \\
    \midrule
    Vanilla Fine-tuning  & 66.10 & 85.20 & 92.62 & 92.04 & \underline{83.99} & 47.42 & 88.23 & 92.95 & 81.86 & \underline{77.62} \\
    \ChildTuningF  & 65.99 & 84.80 & 92.66 & 92.15 & \underline{83.90} & 62.31 & 88.41 & 93.09 & 91.73 & \underline{83.89} \\
    \ChildTuningD  & \bf 66.71 & \bf 86.14 & \bf 92.78 & \bf 92.36 & \bf \underline{84.50} & \bf 70.62 & \bf 88.90 & \bf 93.32 & \bf 92.02 & \underline{\bf 86.22} \\
    \bottomrule
    \end{tabular}
}
\caption{\textbf{Comparison between \ChildTuning and vanilla fine-tuning} applied to four widely used large-scale Pretrained Language Models (PLMs). Average scores on all tasks are \underline{underlined}. The best results are \textbf{bold}. It shows that \ChildTuning yields consistent improvements across all tasks among different PLMs, especially for \ChildTuningD that detects the child network in a task-driven way.
}
\label{table:main}
\end{table*}

%% file: main/experiments.tex
\subsection{Datasets}
\paragraph{GLUE benchmark}
Following previous studies~\citep{mixout, Dodge2020}, we conduct experiments on various datasets from GLUE leaderboard~\citep{wang2018glue}, including linguistic acceptability (CoLA), natural language inference (RTE, QNLI, MNLI), paraphrase and similarity (MRPC, STS-B, QQP), and sentiment classification (SST-2).
CoLA and SST-2 are single-sentence classification tasks and the others are involved with a pair of sentences.
The detailed statistics and metrics are provided in Appendix~\ref{appendix:dataset}.
Following most previous works~\citep{Phang2018,mixout,Dodge2020}, we fine-tune the pretrained model on the training set and directly report results on the dev set using the last checkpoint, since the test results are only accessible by the leaderboard with a limitation of the number of submissions.

\paragraph{NLI datasets} 
In this paper, we also conduct experiments to explore the generalization ability of the fine-tuned model based on several Natural Language Inference (NLI) tasks.
Specifically, we additionally introduce three NLI datasets, i.e., SICK~\citep{SICK}, SNLI~\citep{SNLI} and SciTail~\citep{SCITAIL}. We also report results on the dev set consistent with GLUE.

\subsection{Experiments Setup}
We use the pretrained models and codes provided by HuggingFace\footnote{\url{https://github.com/huggingface/transformers}}~\citep{wolf-etal-2020-transformers}, and follow their default hyperparameter settings unless noted otherwise.
Appendix~\ref{appendix:setup} provides detailed experimental setups (e.g., batch size, training steps, and etc.) for BERT$_{\mathrm{LARGE}}$~\citep{bert},
XLNet$_{\mathrm{LARGE}}$~\cite{xlnet},
RoBERTa$_{\mathrm{LARGE}}$~\cite{roberta},
and ELECTRA$_{\mathrm{LARGE}}$~\cite{electra}.
We report the averaged results over $10$ random seeds.\footnote{Our code is available at \url{https://github.com/alibaba/AliceMind/tree/main/ChildTuning} and \url{https://github.com/PKUnlp-icler/ChildTuning}.}

\input{float/table-nli}
\input{float/figure-transfer-mrpc}
\subsection{Results on GLUE Benchmark}
\label{sec:main-results}
In this section, we show the results of four widely used large PLMs on four GLUE tasks: CoLA, RTE, MRPC, and STS-B, following~\citet{mixout}.
Besides vanilla fine-tuning, we also report the results of two variants of \ChildTuning, including both \ChildTuningF  $\left (p_F={0.2, 0.3, 0.4} \right )$ and \ChildTuningD $\left ( p_D={0.1, 0.2, 0.3} \right )$.

As Table~\ref{table:main} illustrates, \textbf{\ChildTuning outperforms vanilla fine-tuning by a large gain across all the tasks on different PLMs}. 
For instance, \ChildTuning yields an improvement of up to $2.08$ average score on XLNet, and $8.60$ average score on ELECTRA.
Besides, the straightforward task-free variant, \ChildTuningF, can still provide an improvement of $0.87$ average score on BERT and $6.27$ on ELECTRA.
\ChildTuningD, which detects child network in a task-driven way, is more aware of the unique characteristics of the downstream task, and therefore achieves the best performance, with up to $1.50$ and $8.60$ average score improvement on BERT and ELECTRA.
In summary, we can come to a conclusion that \ChildTuning is model-agnostic and can consistently outperform vanilla fine-tuning on different PLMs.

\subsection{Probing Generalization Ability of the Fine-tuned Model}
\label{sec:generalization}
To measure the generalization properties of various fine-tuning methods, in this section, we conduct probing experiments from two aspects, that is, domain generalization and task generalization.

\subsubsection{Domain Generalization}

Besides boosting performance on the target downstream task, we also expect \ChildTuning can help the fine-tuned model achieve better generalization ability towards out-of-domain data.

We evaluate how well the fine-tuned model generalizes to out-of-domain data based on several Natural Language Inference (NLI) tasks.
In detail, we fine-tune BERT$_{\mathrm{LARGE}}$ with different strategies on $5k$ subsampled MNLI and SNLI datasets respectively, and directly test the accuracy of the fine-tuned models on other NLI datasets in different domains, including MNLI, MNLI-mismatch\footnote{MNLI-m has different domain from MNLI training data.}, SNLI, SICK, SciTail, and QQP\footnote{
The target tasks may have different label spaces and we introduce the label mapping in Appendix~\ref{sec:mapping}.
}.
As Table~\ref{table:nli} illustrates, \ChildTuning outperforms vanilla fine-tuning across different out-of-domain datasets.
Specifically, \ChildTuningF improves $1.11$/$0.35$ average score for models trained on MNLI/SNLI, while \ChildTuningD improves up to $1.53$/$0.81$ average score.
In particular, \ChildTuningD achieves $1.90$ score improvement on SICK task and $1.56$ on SNLI task for models trained on MNLI.

The results suggest that \ChildTuning encourages the model to learn more general semantic features during fine-tuning, rather than some superficial features unique to the training data.
Hence, the fine-tuned model can well generalize to different datasets, even though their domains are quite different from the dataset the model is trained on.

\input{float/table-other-methods}
\subsubsection{Task Generalization}
To justify the generalization ability of the model from another perspective, we follow the probing experiments from ~\citet{rxf}, which first freezes the representations from the model trained on one task and then only trains a linear classifier on top of the model for another task.

In particular, we fine-tune BERT$_{\mathrm{LARGE}}$ on MRPC task, and transfer to four other GLUE tasks, i.e., CoLA, STS-B, QNLI, and QQP.
As Figure~\ref{fig:transfer-mrpc} shows, \ChildTuning consistently outperforms vanilla fine-tuning on different transferred tasks.
Compared with vanilla fine-tuning, \ChildTuningF improves $4.58$ average score ($58.95\rightarrow63.53$), while \ChildTuningD even gains up to $7.06$ average score improvement ($58.95\rightarrow66.01$).

In summary, fine-tuning with \ChildTuning gains better performance when the fine-tuned model is transferred to another task, demonstrating that \ChildTuning can maintain more generalizable representations produced by the model than vanilla fine-tuning.

%% file: float/table-nli.tex
\begin{table*}[t]
\centering
\scalebox{0.80}{
    \begin{tabular}{lcccccccccc}
    \toprule
    \multirow{2}{*}{\bf Datasets} & \multicolumn{5}{c}{MNLI} & 
    \multicolumn{5}{c}{SNLI} \\
    \cmidrule(lr){2-6} \cmidrule(lr){7-11}
    ~ & Vanilla & \CTuningF & $\Delta_F$ & \CTuningD & $\Delta_D$ & Vanilla & \CTuningF & $\Delta_F$ & \CTuningD & $\Delta_D$  \\
    \midrule
    MNLI & \underline{75.30} & \underline{75.95} & \underline{+0.65} & \underline{76.61} & \underline{+1.31} & 65.80 & 66.01 & \mycolor{\textbf{+0.21}} & 66.82 & \mycolor{\textbf{+1.02}} \\
    MNLI--m & 76.50 & 77.79 & \mycolor{\textbf{+1.29}} & 77.98 & \mycolor{\textbf{+1.48}} & 67.71 & 67.27 & --0.44 & 68.48 & \mycolor{\textbf{+0.77}} \\
    SNLI & 69.61 & 70.35 & \mycolor{\textbf{+0.74}} & 71.17 & \mycolor{\textbf{+1.56}} & \underline{82.90} & \underline{83.17} & \underline{+0.27} & \underline{83.66} & \underline{+0.76} \\
    SICK & 48.25 & 49.13 & \mycolor{\textbf{+0.88}} & 50.15 & \mycolor{\textbf{+1.90}} & 51.50 & 51.16 & --0.34 & 51.42 & --0.08 \\
    SciTail & 73.65 & 75.42 & \mycolor{\textbf{+1.77}} & 75.08 & \mycolor{\textbf{+1.43}} & 69.35 & 70.74 & \mycolor{\textbf{+1.39}} & 71.10 & \mycolor{\textbf{+1.75}} \\
    QQP & 71.37 & 72.24 & \mycolor{\textbf{+0.87}} & 72.67 & \mycolor{\textbf{+1.30}} & 70.60 & 71.52 & \mycolor{\textbf{+0.92}} & 71.19 & \mycolor{\textbf{+0.59}} \\
    \midrule
    \textbf{Avg$^*$} & 67.88 & 68.99 & \mycolor{\textbf{+1.11}} & 69.41 & \mycolor{\textbf{+1.53}} & 64.99 & 65.34 & \mycolor{\textbf{+0.35}} & 65.80 & \mycolor{\textbf{+0.81}} \\
    \bottomrule
    \end{tabular}
}
\caption{
\textbf{Probing domain generalization}.
The models are trained on MNLI/SNLI and tested on out-of-domain data.
$\Delta_F$ and $\Delta_D$ denotes the improvement of \CTuningF and \CTuningD  compared with vanilla fine-tuning.
Average scores (marked with $^*$) is computed excluding in-domain results (\underline{underlined}).
Positive transfer results are highlighted in \mycolor{\textbf{blue}}.
\ChildTuning can better maintain the out-of-domain generalization ability of the model.
}
\label{table:nli}
\end{table*}

%% file: float/figure-transfer-mrpc.tex
\begin{figure*}[htbp]
\centering
\subfigure[CoLA (Matthews Corr)]{
\begin{minipage}[t]{0.23\linewidth}
\centering
\includegraphics[width=\linewidth]{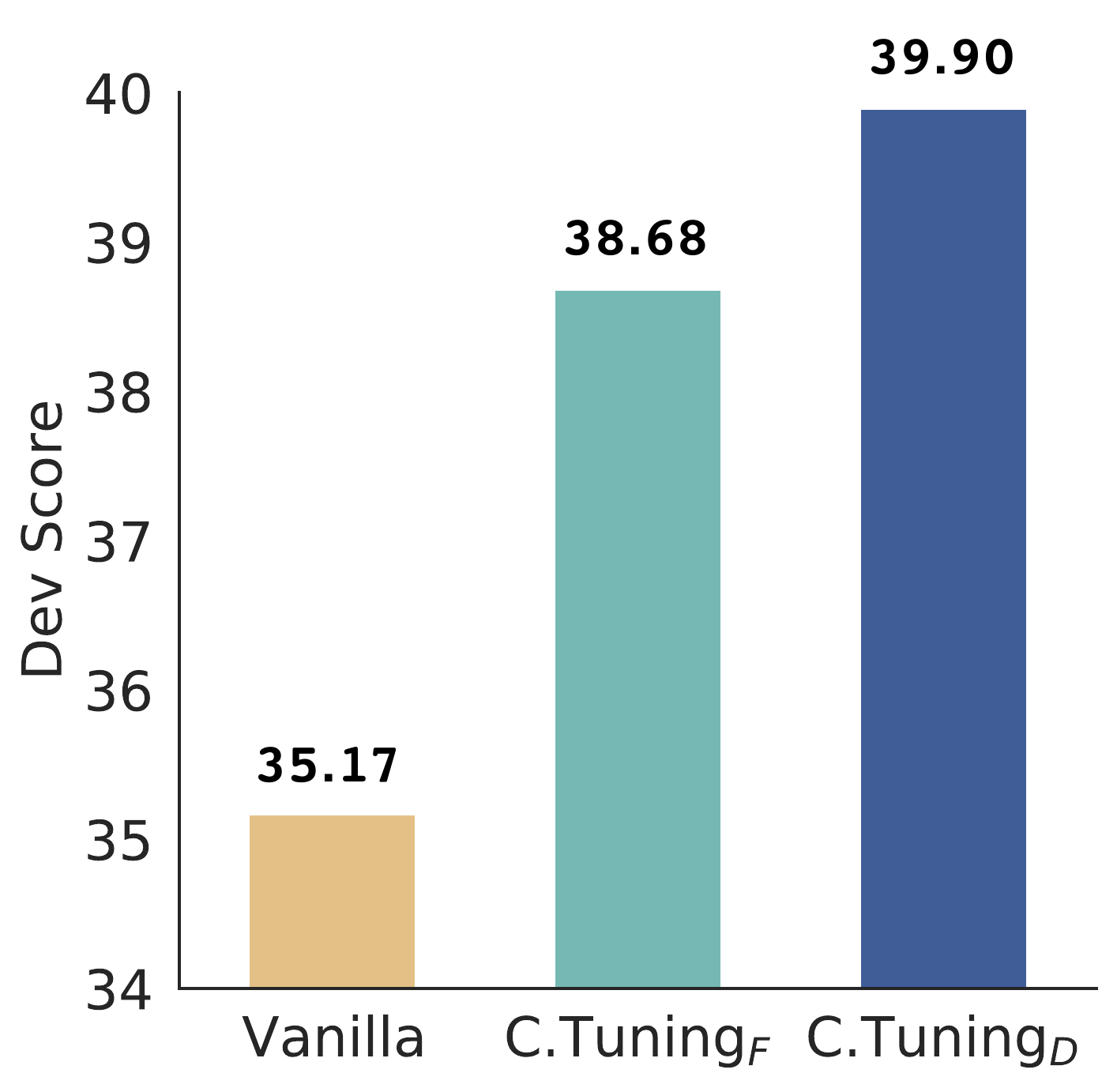}
\end{minipage}
}
\subfigure[STS-B (Spearman Corr)]{
\begin{minipage}[t]{0.23\linewidth}
\centering
\includegraphics[width=\linewidth]{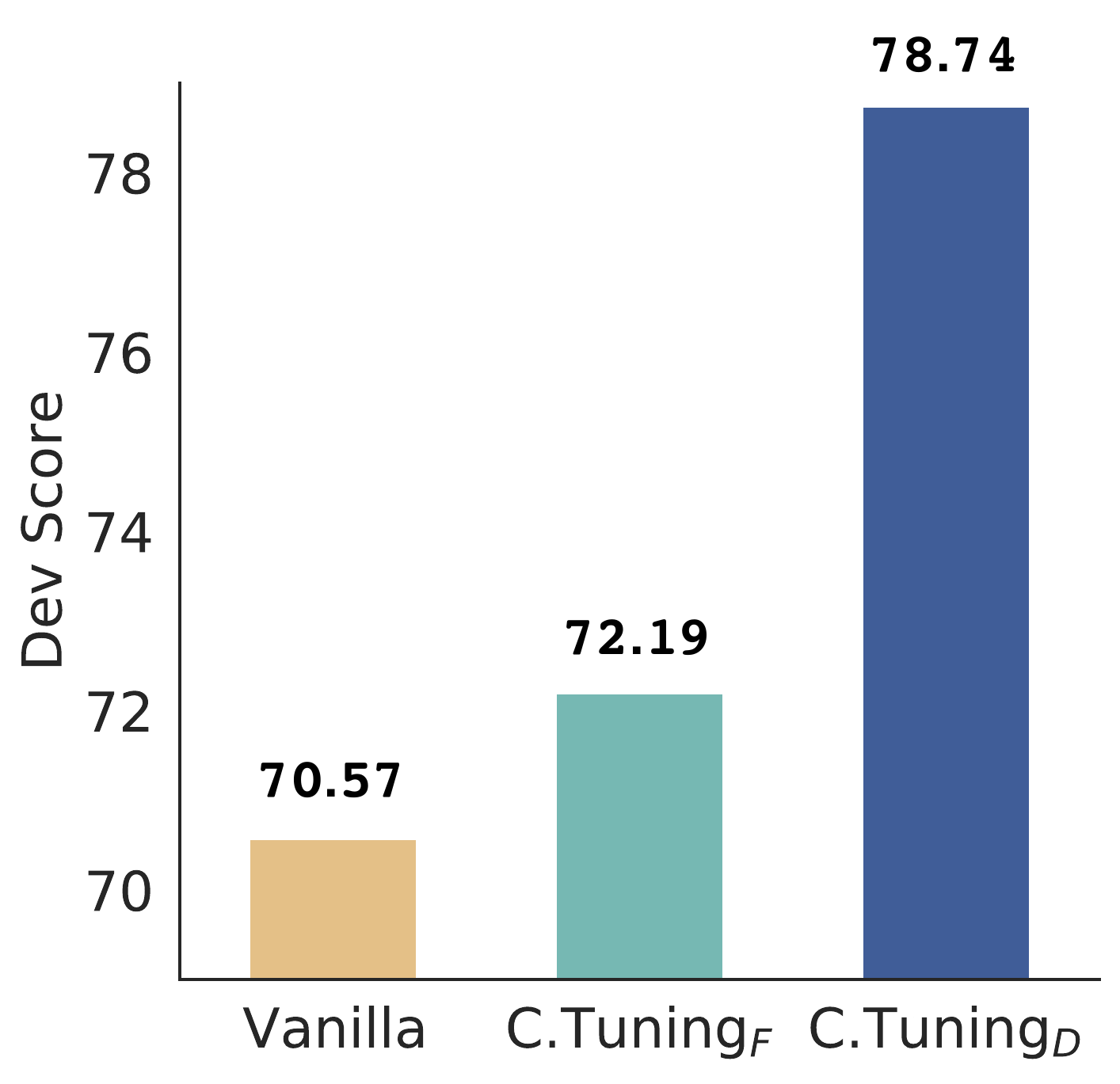}
\end{minipage}
}
\subfigure[QNLI (Accuracy)]{
\begin{minipage}[t]{0.24\linewidth}
\centering
\includegraphics[width=\linewidth]{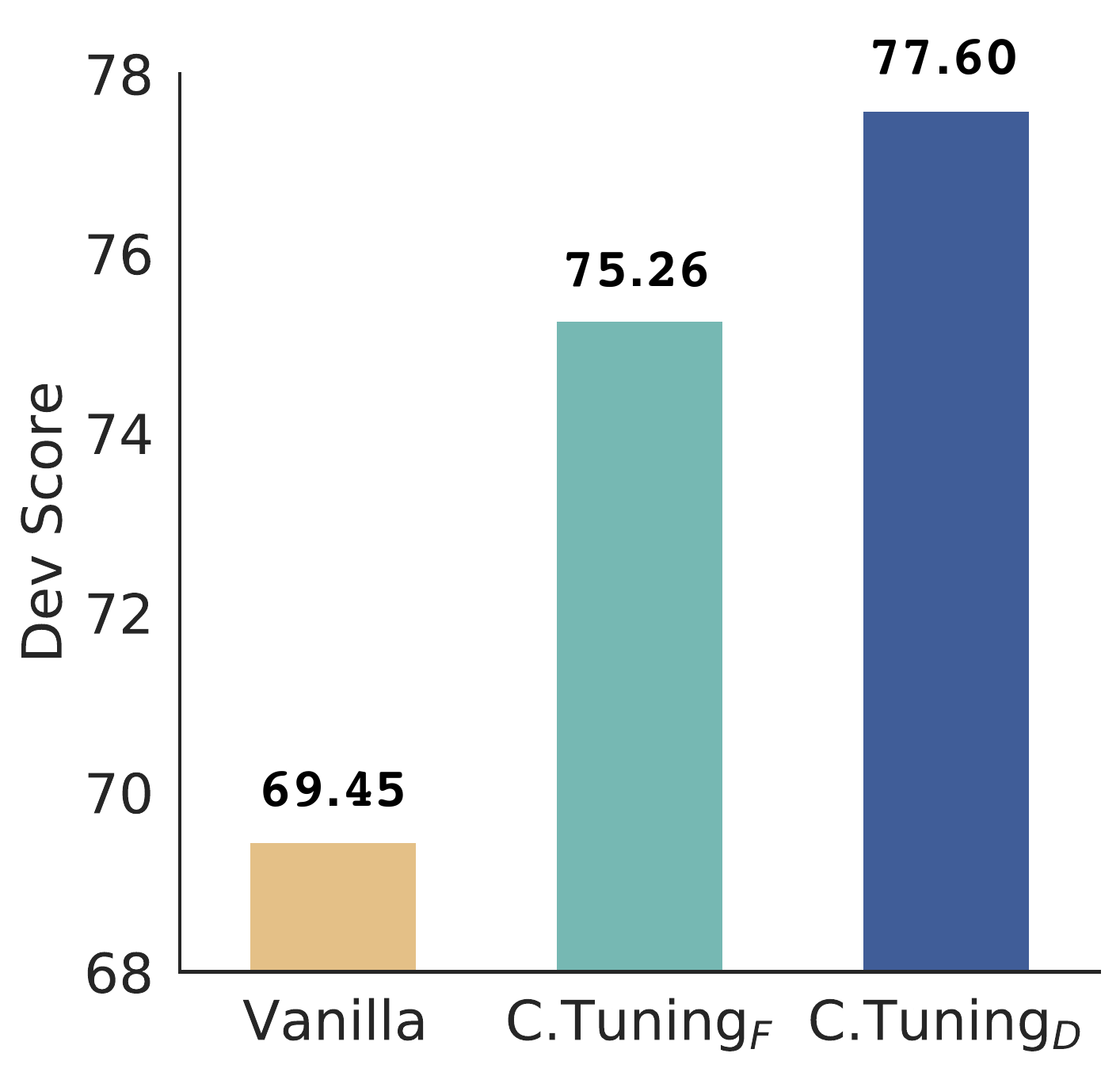}
\end{minipage}
}%
\subfigure[QQP (F1)]{
\begin{minipage}[t]{0.24\linewidth}
\centering
\includegraphics[width=\linewidth]{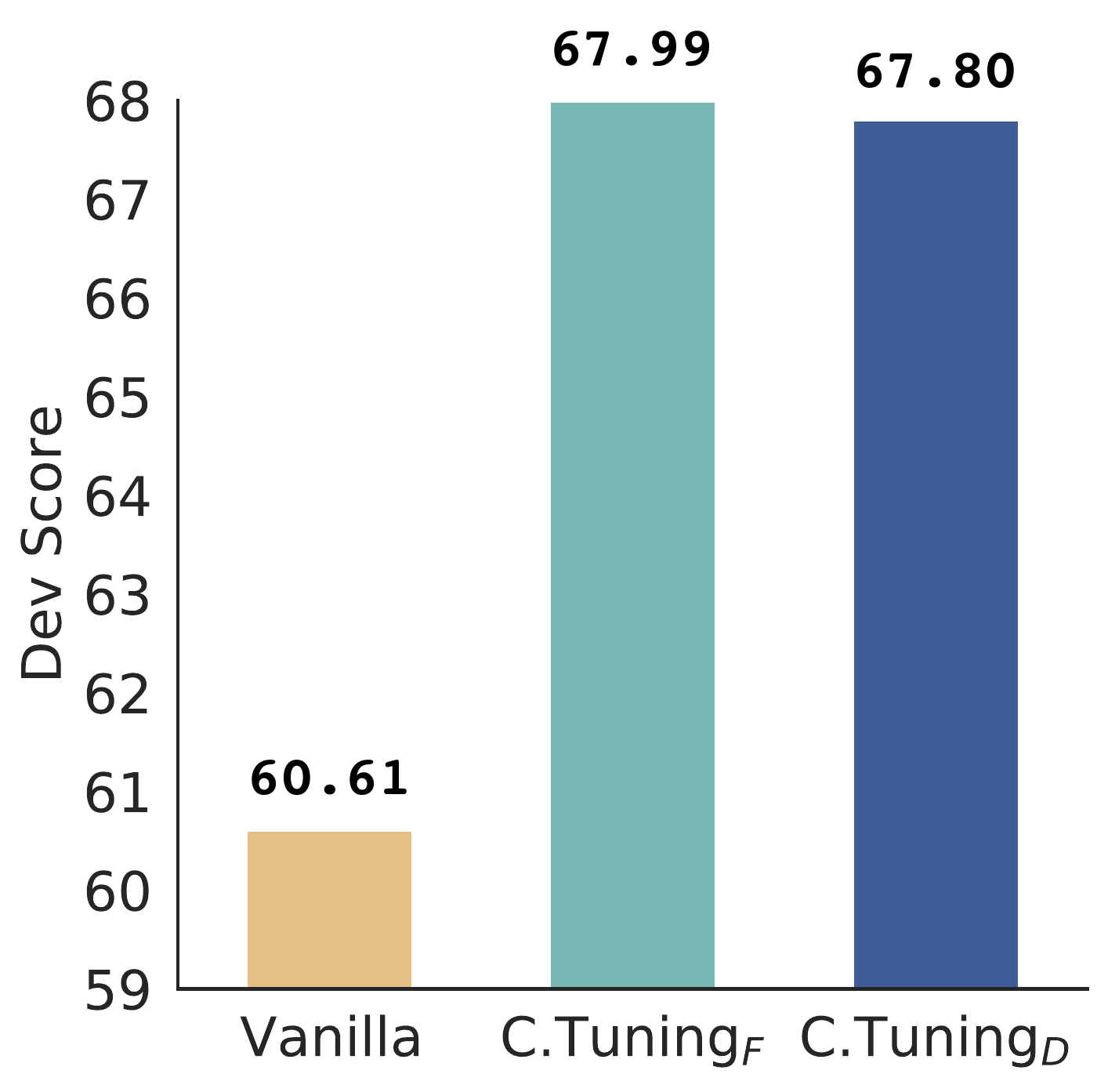}
\end{minipage}
}
\centering
\caption{
\textbf{Probing task generalization}.
The model is fine-tuned on MRPC task and transferred to four different tasks.
\ChildTuning can maintain more generalizable representations compared with vanilla fine-tuning.}
\label{fig:transfer-mrpc}
\end{figure*}

%% file: float/table-other-methods.tex
\begin{table*}[t]
\centering
\scalebox{0.84}{
    \begin{tabular}{lcccccc}
    \toprule
    \bf Methods & CoLA & RTE & MRPC & STS-B  & \bf Avg  & $\Delta$ \\
    \midrule
    Vanilla Fine-tuning$^\dagger$  & 60.60 (~~~--~~~) & 70.40 (~~~--~~~) & 88.00 (~~~--~~~) & 90.00 (~~~--~~~) & 77.25 & ~~~--~~~ \\
    Vanilla Fine-tuning & 63.13 (64.31) & 70.18 (72.56) & 90.77 (91.42) & 89.61 (90.12) & 78.42 & 0.00 \\
    \midrule
    Weight Decay~\cite{daume-iii-2007-frustratingly}  & 63.63 (64.56) & 71.99 (74.37) & 90.93 (91.70) & 89.82 (90.29) & 79.09 & +0.67  \\
    Top-$K$ Tuning~\cite{pmlr-v97-houlsby19a} & 62.63 (64.06) & 70.90 (74.73) & 91.09 (92.20) & 89.97 (90.15) & 78.65 & +0.23 \\
    Mixout~\cite{mixout} & 63.60 (64.82) & 72.15 (75.45) & 91.29 (91.85) & 89.99 (90.13) & 79.26 & +0.84 \\
    RecAdam~\cite{RecAdam} & 64.33 (65.33) & 71.63 (73.29) & 90.85 (92.01) & 89.86 (90.42) & 79.17 & +0.75 \\
    R3F~\cite{rxf} & 64.13 (66.32) & 72.28 (74.73) & 91.18 (91.57) & 89.61 (90.12) & 79.30 & +0.88 \\
    \midrule
    \ChildTuningF & 63.71 (66.06) &  72.06 (74.73) &  91.22 (91.85) &  \bf 90.18 (90.92) & 79.29 &  +0.87 \\
    \ChildTuningD & \bf 64.92 (66.03) & \bf 73.14 (76.17) & \bf 91.42 (92.17) & \bf 90.18 (90.64) & \bf 79.92 & \bf +1.50 \\
    \midrule
    \ChildTuningD + R3F & \bf 65.18 (66.03) & \bf 73.43 (76.17) & \bf 92.23 (92.65) & \bf 90.18 (90.64)$^{*}$ &  \bf 80.26 & \bf +1.84 \\
    \bottomrule
    \end{tabular}
}
\caption{
\textbf{Comparison between \ChildTuning with other fine-tuning methods}.
We report the mean (max) results of $10$ random seeds.
Results with $^\dagger$ are taken from~\citet{xlnet}, and others are from our implementation.
The task-driven variant, \ChildTuningD, achieves the best performance compared with other methods.
Integrating \ChildTuningD with other fine-tuning methods like R3F can yield further improvements.
Note that since R3F is not applicable to regression task, the result on STS-B (marked with $^*$) is the same as \ChildTuningD.
}
\label{table:other-methods}
\end{table*}

%% file: main/discussion.tex
\subsection{Comparison with Prior Methods}
\label{sec:other-methods}
In this section, we review and compare prior studies towards effective fine-tuning:
1) \textbf{Weight Decay}~\citep{daume-iii-2007-frustratingly}, which adds the $\lambda \|\vect{w}-\vect{w}_0 \|_2$ penalty to the loss function, where $\vect{w}_0$ denotes the pretrained weights;
2) \textbf{Top-$K$ Tuning}, which only fine-tune the top-$K$ layers of the model with other layers freezed. ~\citet{pmlr-v97-houlsby19a} uses it as a strong baseline;
3) \textbf{Mixout}~\citep{mixout}, which randomly replaces the parameters with their pretrained weights; 
4) \textbf{RecAdam}~\citep{RecAdam}, which is similar to Weight Decay while its loss weights $\lambda$ keeps changing during fine-tuning;
5) Robust Representations through Regularized Finetuning (\textbf{R3F})~\citep{rxf}, which is rooted in trust region theory. 
Appendix~\ref{appendix:setting-regularization} shows detailed hyperparameter settings.

We compare \ChildTuning with these methods based on BERT$_{\mathrm{LARGE}}$, and report the mean (max) score results in Table~\ref{table:other-methods}, following ~\citet{mixout}.
While all the fine-tuning methods can bring improvements across four different tasks compared with vanilla fine-tuning, \ChildTuning achieves the best performance.
In detail, among prior fine-tuning methods, Mixout and R3F yield the highest improvement with $0.84$ and $0.88$ average score respectively.
\ChildTuningF has performance on par with Mixout and R3F, while \ChildTuningD achieves $1.50$ average score improvement in total.
More importantly, \ChildTuning is flexible and orthogonal to most fine-tuning methods. Thus, integrating \ChildTuning with other methods can further boost the performance. For instance, combining \ChildTuningD with R3F leads to a $1.84$ average score improvement in total.

In short, compared with prior fine-tuning methods, we find that 
1) \ChildTuning is more effective in adapting PLMs to various tasks, especially for the task-driven variant \ChildTuningD, and
2) \ChildTuning has the advantage that it is flexible enough to integrate with other methods to potentially achieve further improvements.

\subsection{Results in Low-resource Scenarios}
\input{float/table-subsample}

Fine-tuning a large pretrained model on extremely small datasets can be very challenging since the risk of overfitting rises~\citep{Dodge2020}.
Thus, in this section, we explore the effect of \ChildTuning with only a few training examples.
To this end, we downsample all datasets in GLUE to $1$k training examples and  fine-tune BERT$_{\mathrm{LARGE}}$ on them.

As Table~\ref{table:subsample} demonstrates, compared with vanilla fine-tuning, \ChildTuningF improves the average score by $1.42$, and the improvement is even larger for \ChildTuningD, which is up to $2.24$.
It suggests that although overfitting is quite severe when the training data is in extreme low-resource scenarios, \ChildTuning can still effectively improve the model performance, especially for \ChildTuningD since it decreases the hypothesis space of the model.

\subsection{What is the Difference Between \ChildTuning and Model Pruning?}

\ChildTuningD detects the most important child network in a task-driven way, and only updates this parameters within the child network during the fine-tuning with other parameters freezed.
It is very likely to be confused with model pruning~\citep{DBLP:conf/iclr/0022KDSG17, prune, Lin2020Dynamic}, which also detects a subnetwork within the model (but then removes the other parameters).

Actually, \ChildTuning and model pruning are different in both the \emph{objectives and methods}.
Regarding objectives, model pruning aims at improving the inference efficiency and maintaining the performance at the same time, while \ChildTuning is proposed to address the overfitting problem and improve the generalization ability for large-scale language models during fine-tuning.
Regrading methods, model pruning abandons the unimportant parameters during inference, while the parameters that do not belong to the child network are still reserved for \ChildTuning during training and inference.
In this way, the knowledge of the non-child network hidden in the pretrained weights will be fully utilized.

To better illustrate the effectiveness of \ChildTuningD compared to model pruning,
we set all the parameters not belonging to the child network to zero, which is referred to as \underline{Prune} in Table~\ref{table:ablation}.
It shows that, once we abandon parameters out of the child network, the score dramatically decreases by $33.89$ points averaged on four tasks (CoLA/RTE/MRPC/STS-B), and the model even collapses on CoLA task.
It also suggests that besides parameters in child network, those in the non-child network are also necessary since they can provide general knowledge learned in pretraining.

\input{float/table-ablation}

\subsection{Is the Task-Driven Child Network Really that Important to the Target Task?}

\ChildTuningD detects the task-specific child network by means of choosing parameters with the highest Fisher information towards the downstream task data.
In this section, we exlore whether the detected task-driven child network is really that important to the task.

To this end, we introduce two ablation studies for \ChildTuningD:
1) \underline{Random}: We randomly choose a child network and keep it unchanged during fine-tuning;
2) \underline{Lowest Info.}: We choose those parameters with \emph{lowest} Fisher information as the child network, contrasted to the \emph{highest} Fisher information adopted in \ChildTuningD.

As shown in Table~\ref{table:ablation}, choosing the child network randomly can even outperform vanilla fine-tuning, with $0.18$ average score improvement.
\textbf{It supports our claim that there is no need to update all parameters of the large PLMs}, and decreasing the hypothesis space can reduce the risk of overfitting.
However, it is still worth finding a proper child network to further boost the performance.
If we choose parameters with the lowest Fisher information (Lowest Fisher), the average score is dramatically decreased by $6.65$ compared with choosing with the highest Fisher information adopted in \ChildTuningD.
Hence, we can conclude that the child network detected by \ChildTuningD is indeed important to the downstream task.

\subsection{What is the Relationship among Child Networks for Different Tasks?}
\input{float/figure-overlap}

As the task-driven child networks are correlated with the tasks, we further explore the relationship among child networks for different tasks.
To this end, we visualize the overlapping rate among different task-driven child networks, where we use the Jaccard similarity coefficient, $\frac{\left |  \mathcal{C}^i \cap  \mathcal{C}^j \right |}{\left |  \mathcal{C}^i \cup  \mathcal{C}^j \right |}$, to calculate the overlapping rate between task $i$ and $j$.

Figure~\ref{fig:overlap} shows the overlap among GLUE tasks. As we expected, similar tasks tend to have higher overlapping ratios of child network.
For example, the overlapping ratio among NLI tasks is remarkably higher than others, such as RTE and QNLI, QNLI and MNLI.
For different kinds of tasks, their overlapping ratio is relatively lower, such as CoLA and MRPC.
It is also interesting to find that the task-driven child network for SST2 overlaps less with other tasks except CoLA, even though SST2 and CoLA is not so similar.
The reason may be that both SST2 and CoLA belongs to a single sentence classification task, while others are in a different format of sentence-pair classification tasks.

%% file: float/table-subsample.tex
\begin{table}[t]
\centering
\scalebox{0.92}{
    \begin{tabular}{cccc}
    \toprule
     \bf Dataset & Vanilla & \CTuningF & \CTuningD  \\
     \midrule
     CoLA & 47.48  & 48.44  & \bf 50.37 \\
     RTE & 65.09   & 65.52 & \bf 68.09 \\
     MRPC & 84.91  & 85.44 & \bf 86.49 \\
     STS-B & 81.86 & 82.25 & \bf 82.76 \\
     SST2 & 90.25 & 90.34 &  \bf 90.39 \\
     QNLI & 81.68  & 83.09 & \bf 83.42 \\
     QQP & 71.30  & \bf 72.15 &  71.79 \\
     MNLI & 55.72 & 62.47 &  \bf 62.93 \\
     \midrule
     \textbf{Avg} & 72.29 & 73.71 & \bf 74.53 \\
    \bottomrule
    \end{tabular}
}
\caption{\textbf{Results in low-resource scenarios}. \ChildTuning is better than vanilla fine-tuning in alleviating overfitting problems.}
\label{table:subsample}
\end{table}

%% file: float/table-ablation.tex
\begin{table}[t]
\centering
\scalebox{0.9}{
    \begin{tabular}{lcccccc}
    \toprule
    \bf Methods & CoLA & RTE & MRPC & STS-B \\
    \midrule
    Vanilla & 63.13 & 70.18 & 90.77 & 89.61 \\
    \midrule
    Prune & 0.00 & 51.12 & 81.40 & 45.63  \\
    Random & 63.23 & 70.69 & 90.83 & 89.67 \\
    Lowest Info. & 60.33 & 59.86 & 83.82 & 88.52 \\
    \midrule
    \CTuningD & \textbf{64.92} &\textbf{ 72.78} & \textbf{91.26} & \textbf{90.18} \\
    \bottomrule
    \end{tabular}
}
\caption{
\textbf{Ablation study of \ChildTuningD}.
\underline{Prune}: Abandon parameters out of the child network.
\underline{Random}: Randomly choose a child network and keep it unchanged during fine-tuning. 
\underline{Lowest Info.}: Detect a child network with lowest Fisher information instead.
}
\label{table:ablation}
\end{table}

%% file: float/figure-overlap.tex
\begin{figure}[t]
	\centering
    \includegraphics[width=0.42\textwidth]{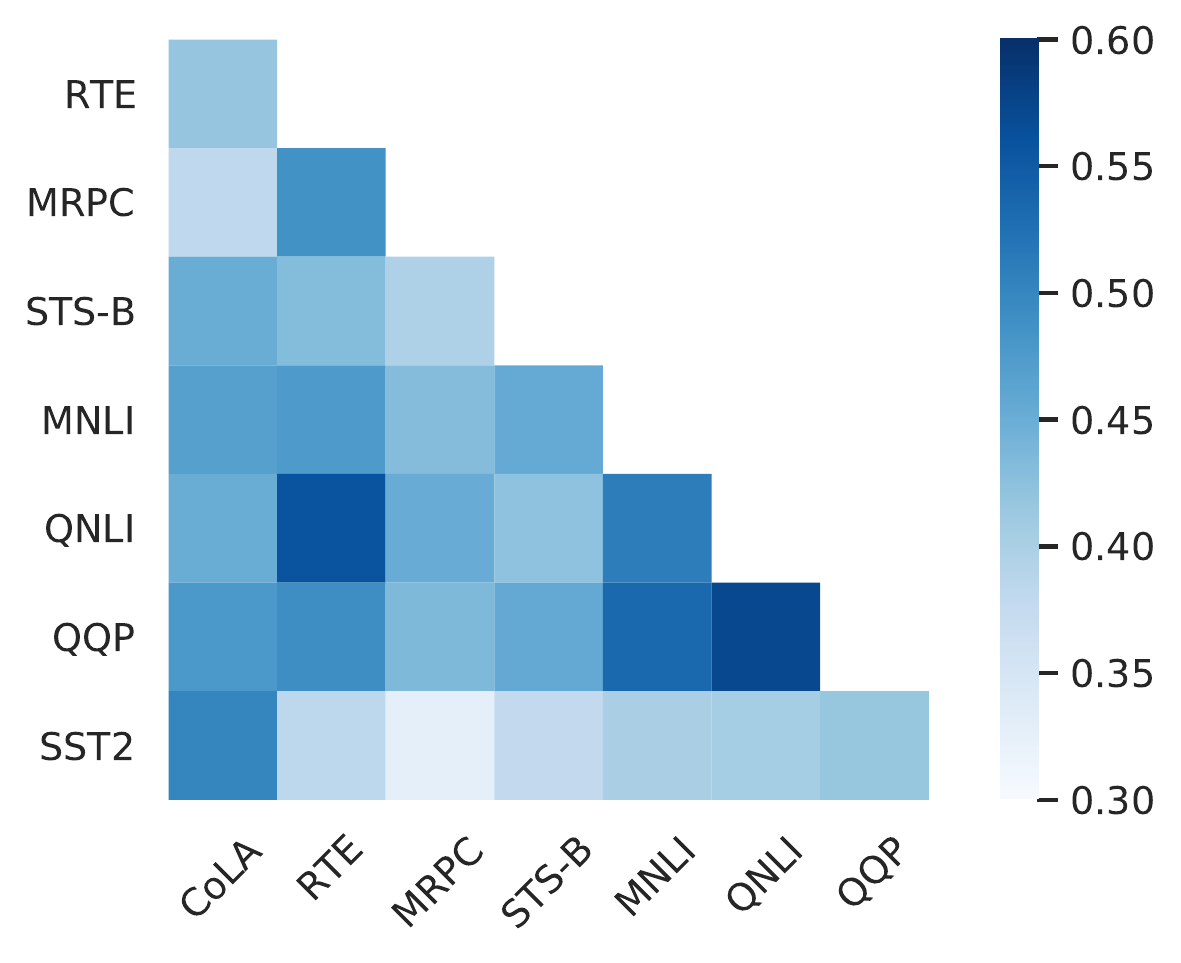}
	\caption{The overlapping ratio among task-driven child networks among GLUE tasks.}
	\label{fig:overlap}
\end{figure}

%% file: main/related.tex
\paragraph{Explosion of PLMs.}
There has been an explosion of studies on Pretrained Language Models (PLMs).
\citet{bert} propose BERT that is pretrained on large quantities of unannotated corpus with self-supervised tasks.
Many PLMs also emerged such as GPT-$2$~\cite{gpt2}, GPT-$3$~\cite{gpt3},
ELECTRA~\citep{electra}, XLNet~\citep{xlnet}, RoBERTa~\citep{roberta}, and BART~\citep{bart}.
The number of parameters of PLMs also explodes.
BERT$_{\mathrm{LARGE}}$ has $340$ millions of parameters, and the number for GPT-$3$ is even up to $175$ billions.

\paragraph{Effective and generalizable fine-tuning.}
With a mass of parameters, fine-tuning large PLMs tend to achieve degenerated performance due to overfitting and have poor generalization ability, especially on small datasets~\citep{bert, Phang2018, mixout}.
Therefore, different fine-tuning techniques have been proposed.
Some of them utilize the pretrained weights to regularize the deviation of the fine-tuned model~\citep{mixout, daume-iii-2007-frustratingly, RecAdam},
while others compress the output information~\citep{VIB} or injects noise into the input~\citep{smart, rxf}.
Moreover, \citet{revisiting} and \citet{Stability20Mosbach} point out that the omission of bias correction in the Adam optimizer used in \citet{bert} is also responsible for the degenerated results.
Orthogonal to these methods, \ChildTuning address the problems by detecting the child network within the model in a task-free or task-driven way. It only updates parameters within the child network via a gradient mask, which is proved to be effective in adapting large PLMs to various tasks, along with better generalization ability.

\paragraph{Parameter-efficient Fine-tuning.}
There are also studies focusing on parameter-efficient fine-tuning, for example, the adapter-based methods~\citep{pmlr-v97-houlsby19a,pfeiffer-etal-2020-mad,karimi-mahabadi-etal-2021-parameter}, and the Diff-Pruning method~\citep{guo-etal-2021-parameter}.
However, our \ChildTuning is different from this line of works.
Firstly, they aim at fine-tuning as few as possible parameters to maintain performance, while we target effective and generalizable fine-tuning. 
Secondly, Diff-Pruning sparsifies diff-vector with gradient estimators, and adapter-based methods fine-tune new added module during training, while we detect the child network inside the model without extra parameters and only need to calculate the FIM before training for \ChildTuningD.
Finally, we consistently outperform vanilla fine-tuning by a large margin, while they achieve competitive performance with full model training.

%% file: main/conclusion.tex
To mitigate the overfitting problem and improve generalization for fine-tuning large-scale PLMs, we propose a straightforward yet effective fine-tuning technique, \ChildTuning, which only updates the child network during fine-tuning via strategically masking out the gradients of the non-child network.
Two variants are introduced, \ChildTuningF and \ChildTuningD, which detect the child network in a task-free and task-driven way, respectively.
Extensive experiments on various downstream tasks show that both of them can outperform vanilla fine-tuning and prior works by large gains among four different pretrained language models, and meanwhile largely enhance the generalization ability of the fine-tuned models.
Since \ChildTuning is orthogonal to most prior fine-tuning techniques, integrating \ChildTuning with them can further boost the performance. 

%% file: main/appendix.tex
\section{GLUE Benchmark Introduction}
\label{appendix:dataset}
In this paper, we conduct experiments on $8$ datasets in GLUE benchmark~\citep{wang2018glue} as shown in Table~\ref{table:glue-benchmark}, including single-sentence tasks, inference tasks, and similarity and paraphrase tasks.
Note that the original GLUE benchmark includes $9$ different datasets in total.
However, there are some issues with the construction of the WNLI dataset\footnote{\url{https://gluebenchmark.com/faq}}.
Therefore most studies exclude this dataset~\citep{bert, Dodge2020} and we follow them.
The metrics we report for each dataset are also illustrated in Table~\ref{table:glue-benchmark}.

\begin{table}[htbp]
\centering
\begin{tabular}{lccc}
\toprule
\bf Dataset & \bf \#Train & \bf \#Dev & \bf Metrics \\
\midrule
\multicolumn{3}{l}{\textit{Single-sentence Tasks}} \\
CoLA & 8.5k & 1.0k & Matthews Corr \\
SST-2 & 67k & 872 & Accuracy \\
\midrule
\multicolumn{3}{l}{\textit{Inference}} \\
RTE & 2.5k & 277 & Accuracy \\
QNLI & 105k & 5.5k & Accuracy \\
MNLI & 393k & 9.8k & Accuracy \\
\midrule
\multicolumn{3}{l}{\textit{Similarity and Paraphrase}} \\
MRPC & 3.7k & 408 & F1 \\
STS-B & 5.7k & 1.5k & Spearman Corr \\
QQP & 364k & 40k & F1 \\
\bottomrule
\end{tabular}
\caption{Statistics and metrics of eight datasets used in this paper form GLUE benchmark.
}
\label{table:glue-benchmark}
\end{table}

\section{Settings for Different Pretrained Language Models}
\label{appendix:setup}
In this paper, we fine-tune different large pretrained language models with \ChildTuning, including
BERT$_{\mathrm{LARGE}}$\footnote{\url{https://huggingface.co/bert-large-cased/tree/main}},
XLNet$_{\mathrm{LARGE}}$\footnote{\url{https://huggingface.co/xlnet-large-cased/tree/main}},
RoBERTa$_{\mathrm{LARGE}}$\footnote{\url{https://huggingface.co/roberta-large/tree/main}}, 
and ELECTRA$_{\mathrm{LARGE}}$\footnote{\url{https://huggingface.co/google/electra-large-discriminator/tree/main}}.
The training epochs/steps, batch size, and warmup steps are listed in Table~\ref{table:settings}.
We use AdamW~\citep{loshchilov2018decoupled} optimizer, and set $\beta_1=0.9$, $\beta_2=0.999$, $\epsilon=1\text{e-}6$.
We clip the gradients with a maximum norm of $1$, and the maximum sequence length is set as $128$.
For \ChildTuningF, we uses $p_F=\left \{ 0.2, 0.3, 0.4\right \}$ and re-scale the gradients to ensure the gradients after \ChildTuningF are unbiased.
For \ChildTuningD, we use $p_D=\left \{ 0.1, 0.2, 0.3\right \}$.
We use grid search for learning rate from $\left \{ 1\text{e-}5, 2\text{e-}5, \dots, 1\text{e-}4 \right \}$.
We conduct all the experiments on a single GTX-3090 GPU.

\begin{table*}[htbp]
\centering
\begin{tabular}{ccccc}
\toprule
\bf Model & \bf Dataset & \bf Batch Size & \bf Training Epochs/Steps & Warmup Ratio/Steps \\
\midrule
BERT & all & 16 & 3 epochs & 10\% \\
\midrule
\multirow{4}{*}{XLNet} & CoLA & 128 & 1200 steps & 120 steps \\
~ & RTE & 32 & 800 steps & 200 steps \\
~ & MRPC & 32 & 800 steps & 200 steps \\
~ & STS-B & 32 & 3000 steps & 500 steps  \\
\midrule
\multirow{4}{*}{RoBERTa} & CoLA & 16 & 5336 steps & 320 steps \\
~ & RTE & 16 & 2036 steps & 122 steps \\
~ & MRPC & 16 & 2296 steps & 137 steps \\
~ & STS-B & 16 & 3598 steps & 214 steps  \\
\midrule
\multirow{4}{*}{ELECTRA} & CoLA & 32 & 3 epochs & 10\% \\
~ & RTE & 32 & 10 epochs & 10\% \\
~ & MRPC & 32 & 3 epochs & 10\% \\
~ & STS-B & 32 & 10 epochs & 10\% \\
\bottomrule
\end{tabular}
\caption{Hyperparameters settings for different pretrained models on variant tasks. These settings are reported in the their official repository for \emph{best practice}.}
\label{table:settings}
\end{table*}

These pretrained models are all Transformer-based.
XLNet~\citep{xlnet} is an autoregressive pretrained language model with token permutations.
It generates tokens in an autoregressive way while can still capture bidirectional context information.
RoBERTa~\citep{roberta} is a robustly optimized version of BERT.
It uses a dynamic masking mechanism, larger batch size, and longer training times, and it also abandons the next sentence prediction task.
ELECTRA~\citep{electra} pretrains the model with a generator and a discriminator.
The discriminator is trained to distinguish whether the token is generated by the generator or the original token.

\section{Settings for Other Fine-tuning Methods}
\label{appendix:setting-regularization}

We compare Child-tuning with several other regularization approaches in our paper.
In this section, we simply introduce these approaches and their hyperparameters settings.

\paragraph{Weight Decay}
\citet{daume-iii-2007-frustratingly} proposes to adds a penalty item to the loss function to regulate the $L_2$ distance between fine-tuned models and the pretrained models.
Therefore, the loss function is as follows:
\begin{align*}
\mathcal{L} (\vect{w}) =  \mathcal{L}_{\mathrm{CE}} (\vect{w}) + \lambda_{\mathrm{WD}}\|\vect{w}-\vect{w}_0 \|_2]
\label{eq:weight-decay}
\end{align*}
We grid search the optimal $\lambda_{\mathrm{WD}}$ from $\left \{ 10, 1, 10^{-1}, 10^{-2}, 10^{-3}, 10^{-4}\right \}$.

\paragraph{Top-$K$ Fine-tuning}
Top-$K$ Fine-tuning is a common method and \citet{pmlr-v97-houlsby19a} uses it as a strong baseline.
Top-$K$ Fine-tuning only updatess the top $K$ layers along with the classification layer, while freezing all the other bottom layers.
We grid search the optimal $K$ from $\left \{ 0, 3, 6, 12\right \}$ in our paper.

\paragraph{Mixout}
~\citet{mixout} randomly replace the parameters with its pretrained weights with a certainly probability $p$ during fine-tuning, which aims to minimize the deviation of the fine-tuned model towards the pretrained weights.
In our paper, we grid search the optimal $p$ from $\left \{ 0.1, 0.2, \dots, 0.8\right \}$.
We use the implementation in \url{https://github.com/bloodwass/mixout}.

\paragraph{RecAdam}
~\citet{RecAdam} proposes a new optimizer RecAdam for fine-tuning, which can be considered as an advanced version of Weight Decay, because the coefficient of two different loss items are changed as the training progresses.
The following equations demonstrate the new loss function, where $k$ and $t_0$ are controlling hyperparameters and $t$ is the current training step.
\begin{align*}
\mathcal{L} (\vect{w}) &=  \lambda_{\mathrm{Rec}}(t)  \mathcal{L}_{\mathrm{CE}} (\vect{w}) \\ 
& + (1-\lambda_{\mathrm{Rec}}(t))  \|\vect{w}-\vect{w}_0 \|_2
\end{align*}
\begin{align*}
\lambda_{\mathrm{Rec}}(t) = \frac{1}{1+\exp (-k\cdot(t-t_0))} 
\end{align*}
We grid search the $k$ from $\left \{ 0.05, 0.1, 0.2, 0.5, 1.0 \right \}$, and $t_0$
 from $\left \{ 50, 100, 250, 500 \right \}$.
We use the implementation in \url{https://github.com/Sanyuan-Chen/RecAdam}.

\paragraph{Robust Representations through Regularized Fine-tuning (R3F)}
~\citet{rxf} propose R3F for fine-tuning based on trust region theory, which adds noise into the sequence input embedding and tries to minimize the symmetrical KL divergence between probability distributions given original input and noisy input.
The loss function of R3F is as follows:
\begin{align*}
\mathcal{L} (\vect{w}) &=  \mathcal{L}_{\mathrm{CE}} (\vect{w}) + \lambda_{\mathrm{R3F}} KL_S (f(x) || f(x+z))\\
s.t. & \quad z \sim \mathcal{N}(0, \sigma^2I) \quad or \quad z \sim \mathcal{U}(-\sigma, \sigma)
\end{align*}
where $f(\cdot)$ denotes the model and $z$ denotes the noise sampled from either normal distribution or uniform distribution controlled by hyperparameter $\sigma$, and $KL_S(x||y) = KL(x||y) + KL(y||x)$.
We use both normal and unform distribution, $\lambda_{\mathrm{R3F}}=1$, and grid search the $\sigma$ from $\left \{ 0.1, 0.5, 1.0, 5.0 \right \}$.
We use the implementation in \url{https://github.com/pytorch/fairseq/tree/master/examples/rxf}.

\section{Label Mapping in Task Generalization }
\label{sec:mapping}

MNLI and SNLI datasets contain three labels, i.e., \textit{entailment}, \textit{neutral}, and \textit{contradiction}.
For SciTail, it only has two labels, \textit{entailment} and \textit{neutral}, and therefore we map both \textit{neutral} and \textit{contradiction} in source label space to \textit{neutral} in target label space following~\citet{VIB}. 
For QQP, it has two labels, \textit{duplicate} and \textit{not duplicate}, and ~\citet{gong2018natural} interpret them as \textit{entailment} and \textit{neutral} respectively.
We follow ~\citet{gong2018natural} and use the same mapping strategy as SciTail.

%% file: main/appendix-theory.tex
\section{Theoretical Details}
\label{sec:theory}

We theoretically justify the effectiveness of \ChildTuningF.
Assume \ChildTuningF reserves gradients with probability $p_F \in \left (0, 1 \right ] $, and we simply use $p$ to denote $p_F$ in the following content.
Theorem~\ref{thm:appendix-childtuningf} shows the variance of gradients is a strictly decreasing function of $p$.
When $p=1$, it degenerates into normal fine-tuning methods.
Therefore, \ChildTuningF can improve the variance of the gradients of the model.
Next, Theorem~\ref{thm:appendix-childtuningf2} shows that with higher variance, the model can converge to more flat local minima (smaller $\rho$ in Theorem~\ref{thm:appendix-childtuningf2}).
Inspired by studies that show flat minima tends to generalize better~\citep{KeskarMNST17,abs-2006-05620, foret2021sharpnessaware}, we can further prove \ChildTuningF decreases the generalization error bound.

\begin{thm1}
\label{thm:appendix-childtuningf}
Suppose $\mathcal{L}$ denotes the loss function on the parameter $\vect{w}$, for multiple data instances in the training set $\vect{x}\sim \mathcal{S}$, the gradients obey a Gaussian distribution $\mathcal{N}(\frac{\partial \mathcal{L}}{\partial \vect{w}}, \sigma^2_\vect{g}\mathbf{I}_k)$. For a randomly sampled batch $\mathcal{B}\sim \mathcal{S}$, when the learning algorithm is SGD with learning rate $\eta$, the reserving probability of the \ChildTuningF is $p$, then the mean and covariance of the update $\vect{\Delta w}$ are,
\begin{align}
\mathbb{E}[\vect{\Delta w}]&=-\eta \frac{\partial \mathcal{L}}{\partial \vect{w}}\\
\Sigma[\vect{\Delta w}]&=\frac{\eta^2\sigma_\vect{g}^2\mathbf{I}_k}{p|\mathcal{B}|}+
\frac{(1-p)\eta^2\text{diag}\{\frac{\partial \mathcal{L}}{\partial \vect{w}} \}^2}{p}
\label{eq:update}
\end{align}
where $\Sigma$ is the covariance matrix and diag$(\vect{x})$ is the diagonal matrix of the vector $\vect{x}$. 

Specially, when $\vect{w}$ is a local minima, $\mathbb{E}[\vect{\Delta w}]=\vect{0}_k, \Sigma[\vect{\Delta w}]=\sigma^2\mathbf{I}_k$ and $\sigma^2=\frac{\eta^2\sigma_\vect{g}^2}{p|\mathcal{B}|}$ is a strictly decreasing function of $p$.
\end{thm1}

\begin{thm1}
\label{thm:appendix-childtuningf2}
Suppose $\mathcal{L}$ denotes the expected error rate loss function; $\vect{w}_0$ denotes the pretrained parameter; $k$ is the number of parameters; $\vect{w}$ denotes the local minima the algorithm converges to; $\mathbf{H}$ is the Hessian matrix on $\vect{w}$ and $\rho$ is its greatest eigenvalue; $F_k$ is the cumulative distribution function of the $\chi^2(k)$ distribution.

If the next update of the algorithm $\vect{\Delta w}\sim N(\vect{0}_k, \sigma^2\mathbf{I}_k)$ and the training loss increases more than $\epsilon$ with probability  $\delta$, we assume the algorithm will escape the local minima $\vect{w}$. When the following bound holds, the algorithm can converge to the local minima $\vect{w}$, with higher order infinity omitted,
\begin{align}
\rho\le\frac{2\epsilon}{F^{-1}_k(1-\delta)\sigma^2}
\label{eq:rho}
\end{align} 

Suppose the prior over parameters after training is $P=N(\vect{w}_0, \sigma_0^2\mathbf{I}_k)$, the following generalization error bound holds with probability 1-$\delta$ over the choice of training set $\mathcal{S}\sim\mathcal{D}$,
\begin{align}
\text{bound}(\vect{w}) \le\frac{(k\sigma_0^2-\|\vect{w}-\vect{w}_0\|^2)\epsilon}{k F_k^{-1}(1-\delta)\sigma^2}+\mathcal{R}
\label{eq:bound}
\end{align} 
where $\text{bound}(\vect{w})=\mathcal{L}_\mathcal{S}(\vect{w})-\mathcal{L}_\mathcal{D}(\vect{w})$,
$\mathcal{R} =\sqrt{\frac{k\log \left (1+\frac{k\|\vect{w}-\vect{w}_0\|^2_2}{ k\sigma_0^2-{\|\vect{w}-\vect{w}_0\|^2}}\big(1+\sqrt{\frac{\log|\mathcal{S}|}{k}}\big)^2 \right )+4\log\frac{|\mathcal{S}|}{\delta}}{2(|\mathcal{S}|-1)}}$, with higher order infinity omitted.
\end{thm1}

\subsection{Proof of Theorem~\ref{thm:appendix-childtuningf}}

\begin{proof}
Suppose $\vect{g}^{(i)}$ is the gradient of data instance $\vect{x}^{(i)}, (1\le i\le |\mathcal{B}|)$, then $\vect{g}^{(i)}\sim N(\frac{\partial \mathcal{L}}{\partial \vect{w}}, \sigma_\vect{g}^2\mathbf{I}_k)$. Then, define $\vect{g}=\sum\limits_{i=1}^{|\mathcal{B}|}\frac{\vect{g}^{(i)}}{|\mathcal{B}|}$, we have
\begin{align}
\vect{\Delta w}=-\eta\sum\limits_{i=1}^{|\mathcal{B}|}\frac{\vect{g}^{(i)}}{|\mathcal{B}|}\odot M=-\eta\vect{g}\odot M
\end{align} 
Consider $\vect{g}$, we have
\begin{align}
\mathbb{E}[\vect{g}]=\frac{\mathcal{\partial \mathcal{L}}}{\partial \vect{w}},
\Sigma[\vect{g}]=\frac{\sigma_\vect{g}^2\mathbf{I}_k}{\mathcal{|B|}}
\end{align} 

Suppose $\vect{\hat g}=\frac{\vect{g}}{p}\odot M$, therefore,
\begin{align}
\mathbb{E}[\vect{\hat g}]&=\frac{p}{p}\times \frac{\mathcal{\partial \mathcal{L}}}{\partial \vect{w}}=\frac{\mathcal{\partial \mathcal{L}}}{\partial \vect{w}}
\end{align}

Suppose $\hat g_i, g_i$ are the $i$-th dimension of $\vect{\hat g}, \vect{g}$, we have
\begin{align}
\textbf{D}[\hat g_i]&=
\mathbb{E}[\hat g_i^2]-
(\mathbb{E}[\hat g_i])^2\\
&=p\mathbb{E}[(\frac{g_i}{p})^2]-
(\mathbb{E}[\hat g_i])^2\\
&=\frac{\mathbb{E}[g_i^2]}{p}-
(\mathbb{E}[\hat g_i])^2\\
&=\frac{(\mathbb{E}[g_i])^2+\textbf{D}[g_i]}{p}-
(\mathbb{E}[\hat g_i])^2\\
&=\frac{\textbf{D}[g_i]}{p}+
\frac{(1-p)(\mathbb{E}[\hat g_i])^2}{p}
\end{align}

Therefore,
\begin{align}
\Sigma[\vect{\hat g}]=\frac{\sigma_\vect{g}^2\mathbf{I}_k}{p|\mathcal{B}|}+
\frac{(1-p)\text{diag}\{\mathbb{E}[\vect{g}]\}^2}{p}
\end{align}

Therefore, 
\begin{align}
\mathbb{E}[\vect{\Delta w}]&=-\eta \frac{\partial \mathcal{L}}{\partial \vect{w}}\\
\Sigma[\vect{\Delta w}]&=\frac{\eta^2\sigma_\vect{g}^2\mathbf{I}_k}{p|\mathcal{B}|}+
\frac{(1-p)\eta^2\text{diag}\{\frac{\partial \mathcal{L}}{\partial \vect{w}} \}^2}{p}
\label{eq:update}
\end{align}

Specially, when $\vect{w}$ is a local minima, $\frac{\partial \mathcal{L}}{\partial \vect{w}}=\vect{0}_k$. Therefore, $\mathbb{E}[\vect{\Delta w}]=\vect{0}_k, \Sigma[\vect{\Delta w}]=\sigma^2\mathbf{I}_k$ and $\sigma^2=\frac{\eta^2\sigma_\vect{g}^2}{p|\mathcal{B}|}$ is a strictly decreasing function of $p$.
\end{proof}

\subsection{Proof of Theorem~\ref{thm:appendix-childtuningf2}}

\begin{proof}

We first prove Eq.~\ref{eq:rho}. Apply a Taylor expansion on training loss $\mathcal{L}$, notice that $\nabla_\vect{w}\mathcal{L}(\vect{w})=\vect{0}_k$ since $\vect{w}$ is a local minima. When the algorithm can escape the local minima $\vect{w}$, with higher order infinity omitted, we have,
\begin{align}
\epsilon & \le\mathcal{L}(\vect{w}+\vect{v}) - \mathcal{L}(\vect{w}) \\
=& \vect{v}^\text{T}\nabla_\vect{w}\mathcal{L}(\vect{w})+\frac{1}{2}\vect{v}^\text{T}\mathbf{H}\vect{v}+o(\|\vect{v}\|_2^2) \\
\le & \frac{\rho\|\vect{v}\|_2^2}{2}+o(\|\vect{v}\|_2^2) = \frac{\rho\|\vect{v}\|_2^2}{2}
\end{align}

If the probability of escaping, $P_\text{esc}$, we have
\begin{align}
P_\text{esc} &= P(\mathcal{L}(\vect{w}+\vect{\Delta w}) - \mathcal{L}(\vect{w})\ge \epsilon)\\
&\le P(\frac{\rho \|\vect{\Delta w}\|_2^2}{2} \ge \epsilon) \\
&= P( \|\frac{\vect{\Delta w}}{\sigma}\|_2^2 \ge \frac{2\epsilon}{\rho\sigma^2}) 
\end{align}
namely, $P( \|\frac{\vect{\Delta w}}{\sigma}\|_2^2 \le \frac{2\epsilon}{\rho\sigma^2})\le 1-P_\text{esc}$.

Since $\frac{\vect{\Delta w}}{\sigma}\sim N(\vect{0}_k,\mathbf{I}_k)$, $\|\frac{\vect{\Delta w}}{\sigma}\|_2^2\sim \chi^2(k)$, we have,
\begin{align}
P( \|\frac{\vect{\Delta w}}{\sigma}\|_2^2 \le \frac{2\epsilon}{\rho\sigma^2})=F_k(\frac{2\epsilon}{\rho\sigma^2})
\end{align}
when Eq.~\ref{eq:rho} holds,
\begin{align}
P( \|\frac{\vect{\Delta w}}{\sigma}\|_2^2 \le \frac{2\epsilon}{\rho\sigma^2})=F_k(\frac{2\epsilon}{\rho\sigma^2})\\
\ge F_k(F^{-1}_k(1-\delta))=1-\delta
\end{align}

Therefore, $P_\text{esc}\le 1-P( \|\frac{\vect{\Delta w}}{\sigma}\|_2^2\le \frac{2\epsilon}{\rho\sigma^2})\le \delta$. The algorithm will not escape the local minima $\vect{w}$ and can converge to the local minima $\vect{w}$.

To prove Eq.~\ref{eq:bound}, we introduce Lemma~\ref{lemma:Bayes} in paper~\citet{foret2021sharpnessaware}, which is Theorem 2 in the paper.

\begin{lem}
Suppose $d>0$, the prior over parameters is $P=N(\vect{w}_P, \sigma_P^2\mathbf{I}_k)$ and $\sigma_P^2=d^2+\frac{{\|\vect{w}-\vect{w}_P\|^2}}{k}$, the following bound holds with probability  1-$\delta$ over the choice of training set $\mathcal{S}\sim\mathcal{D}$,
\begin{align}
\mathcal{L}_\mathcal{D}(\vect{w})\le
\max\limits_{\|\vect{\Delta w}\|_2\le d}\mathcal{L}_\mathcal{S}(\vect{w}+\vect{\Delta w})+\mathcal{R}
\end{align}
where $k$ denotes the number of parameters and $\mathcal{R}=\sqrt{\frac{k\log \left (1+\frac{\|\vect{w}-\vect{w}_P\|^2_2}{d^2}\big(1+\sqrt{\frac{\log|\mathcal{S}|}{k}}\big)^2 \right )+4\log\frac{|\mathcal{S}|}{\delta}}{2(|\mathcal{S}|-1)}}$, with higher order infinity omitted.
\label{lemma:Bayes}
\end{lem}

$\ $

In Lemma~\ref{lemma:Bayes}, when we set  $\vect{w}_P = \vect{w}_0$ and $\sigma_P=\sigma_0$, we have
$d^2 = \sigma_0^2-\frac{\|\vect{w}-\vect{w}_0\|^2}{k}$ and $\mathcal{R} = \sqrt{\frac{k\log \left (1+\frac{k\|\vect{w}-\vect{w}_0\|^2_2}{ k\sigma_0^2-{\|\vect{w}-\vect{w}_0\|^2}}\big(1+\sqrt{\frac{\log|\mathcal{S}|}{k}}\big)^2 \right )+4\log\frac{|\mathcal{S}|}{\delta}}{2(|\mathcal{S}|-1)}}$.

$\ $

$\ $

With higher order infinity omitted, we have
\begin{align}
&\max\limits_{\|\vect{\Delta w}\|_2\le d}\mathcal{L}_\mathcal{S}(\vect{w}+\vect{\Delta w})=\mathcal{L}_\mathcal{S}(\vect{w})+\frac{\rho d^2}{2}\\
&\le\frac{(k\sigma_0^2-\|\vect{w}-\vect{w}_0\|^2)\epsilon}{k F_k^{-1}(1-\delta)\sigma^2}
\end{align}

Therefore, the following generalization error bound holds,
\begin{align}
\text{bound}(\vect{w})\le\frac{(k\sigma_0^2-\|\vect{w}-\vect{w}_0\|^2)\epsilon}{k F_k^{-1}(1-\delta)\sigma^2}+\mathcal{R}
\end{align}
where higher order infinity is omitted and $\mathcal{R} = \sqrt{\frac{k\log \left (1+\frac{k\|\vect{w}-\vect{w}_0\|^2_2}{ k\sigma_0^2-{\|\vect{w}-\vect{w}_0\|^2}}\big(1+\sqrt{\frac{\log|\mathcal{S}|}{k}}\big)^2 \right )+4\log\frac{|\mathcal{S}|}{\delta}}{2(|\mathcal{S}|-1)}}$. 
\end{proof}